*Article*

# Autonomous bot with ML-based reactive navigation for indoor environment

**Yash Srivastava [1], Saumya Singh [2] and S.P. Syed Ibrahim [3,\***

[1] Undergraduate Student, School of Electronics and Computer Engineering; yashsrivastava.2018a@vitstudent.ac.in
[2] Undergraduate Student, School of Electronics and Computer Engineering; saumyadhananjay.singh2018@vitstudent.ac.in
[3] Professor, School of Computer Science and Engineering; syedibrahim.sp@vit.ac.in
\* Correspondence: syedibrahim.sp@vit.ac.in; Tel.: +91 9944416392
Vellore Institute of Technology, Chennai, India

**Abstract:** Local or reactive navigation is essential for autonomous mobile robots which operate in an indoor environment. Techniques such as SLAM, computer vision require significant computational power which increases cost. Similarly, using rudimentary methods makes the robot susceptible to inconsistent behavior. This paper aims to develop a robot that balances cost and accuracy by using machine learning to predict the best obstacle avoidance move based on distance inputs from four ultrasonic sensors that are strategically mounted on the front, front-left, front-right, and back of the robot. The underlying hardware consists of an Arduino Uno and a Raspberry Pi 3B. The machine learning model is first trained on the data collected by the robot. Then the Arduino continuously polls the sensors and calculates the distance values, and in case of critical need for avoidance, a suitable maneuver is made by the Arduino. In other scenarios, sensor data is sent to the Raspberry Pi using a USB connection and the machine learning model generates the best move for navigation, which is sent to the Arduino for driving motors accordingly. The system is mounted on a 2-WD robot chassis and tested in a cluttered indoor setting with most impressive results.

**Keywords:** autonomous mobile robot; reactive navigation; machine learning; obstacle avoidance; raspberry pi; arduino

## 1. Introduction

Robotics is an interesting and rapidly evolving field. The application of robotics as a branch of engineering expands with the development of technology. With technological advancement, there is a need for such bots in industries where heavy components are required to be moved from one place to the other, or maybe the most recent is the self-driving car, where we expect zero collisions since a single collision could lead to unforeseen circumstance and an unrecoverable loss.

An autonomous mobile robot that can avoid collisions with unexpected obstacles is called an obstacle avoidance robot. In our obstacle avoidance robot, we have used four ultrasonic sensors of which one is placed in the back and three are placed in the front but at different angles. This bot has an Arduino interfaced with these sensors, and a Raspberry Pi to implement machine learning. Generally, the bot is hardcoded, i.e., the if-else statement decides the bot's path depending on the condition. But in our bot, we have added machine learning for deciding the path. Machine learning with robotics is a very new field with a lot to discover and implement [3, 6, 18, 19, 22]. Machines are already making our life easier and giving them intelligence will be a boon for the industry. Many machine learning algorithms are being combined with traditional ones to improve the efficiency in fields like surgical robots [16], robot soccer [4, 7, 8], wall-following robot [9, 12] and robot navigation task [11]. Fumebot [13] proposed a new method using convolutional neural networks which implemented AlexNet for home monitoring robots. This interaction between machine learning with instantaneous robotics-derived data is a rapidly evolving field with the objective to efficiently facilitate the timely delivery of meaningful feedback.

There are many kinds of navigation technologies for mobile robots and drones, such as route planning [1], positioning, and map interpretation as stated by Manh-Cuong Le and My-Ha Le [2], S. Islam and A. Razi in their path planning algorithm where they implemented reinforcement learning with a new method to incorporate the state space for the next location [5]. G. Prabhakar, B. Kailath, S. Natarajan and R. Kumar implemented a high-speed autonomous driving system to detect and classify the obstacle for tracking and positioning [10]. Aditya Kumar Jain [20] proposed a working model of a self-driving car similar to the obstacle avoidance robot, where the path navigation was done by a convolutional neural network which was fed input images by the camera, and output was controlled by the Arduino. Mohamed El-Shamouty et al [17] demonstrated the simulation-driven ML framework on an industrial use case of assembling electrical cabinets addressing the major challenges between simulation to real-world application. Stelian-Emilian Oltean [21] proposed a low-cost mobile robot platform that can move in 2D environments like a line follower robot with other features such as mapping, navigation, etc. It also has a robotic arm attached with one degree of freedom for lifting and transporting obstacles. Similar to [20] Alfa Rossi et al [23] proposed a real-time lane detection and motion planning vehicle that incorporated Pi camera 1.3 to capture which is then processed by Raspberry-Pi 3 Model B. The output of the image processing algorithm is fed as input to Arduino Uno which controls the motor of the vehicle.

**2. Materials and Methods**

*2.1 Components*

HC-SR04 ultrasonic sensors present a low-cost, visibility-independent solution for distance measurement in the range of 5cm-450cm. The project is built on a 2-wheel drive robot chassis, with a L293D motor driver to combine a 6V power supply with HIGH/LOW logic from the Arduino to drive the two bo-motors. A Raspberry Pi 3B is present as an onboard computer on the bot, for executing the machine learning model. The Raspberry Pi is powered by a power bank capable of supplying 5V at 2.1A. A tunnel SSH connection is set up between Raspberry Pi and the local PC to make communication across networks possible.

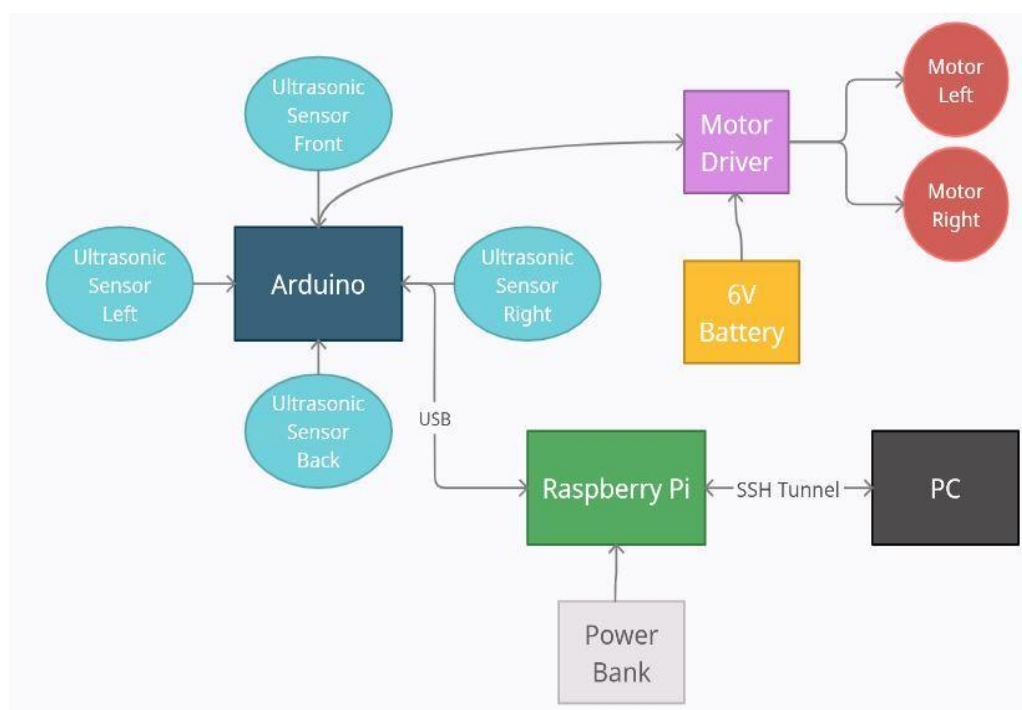

**Figure 1.** Hardware block diagram

*2.2 Software*
2.2.1 Machine learning

There are three types of machine learning: supervised, unsupervised and reinforcement learning. Supervised Learning is used when labelled targets are available. The model learns on the training data and then deployed to be used on the final dataset. Unsupervised Learning is used when labelled targets are not available. The model would have learned some patterns from the training data, but it could find new hidden patterns with the final dataset as well. Reinforcement Learning is used when the model should learn from the new situations. Here, the output from every iteration is fed back into the model and used as the training data. This project already has labelled data, so it falls under supervised machine learning. Thus, supervised machine learning is used to classify the path i.e., whether the robot should move front, back, left, right or stop. Reinforcement learning could also have been used but it would lead to computationally expensive interaction between the robot hardware and the environment. Another disadvantage of using reinforcement learning is that if there is an outlier in the data, the model would be trained with the wrong input, thus affecting the path navigation of the robot [14]. There are different algorithms for classification but the following ones were used and tree-based algorithms returned best accuracy.

- SVM
  SVM or Support Vector Machine is a supervised machine learning algorithm that incorporates the concept of hyperplanes that differentiates between different categories. New data is categorized according to the hyperplanes and their calculation. The accuracy of the SVM classifier on training data was 0.96 and on test data was 0.74.

- KNN

KNN or K-Nearest neighbor is a supervised machine learning algorithm that calculates the similarity between the new data and the available dataset. It then puts the new data into that category that has the maximum similarity. The accuracy of the KNN classifier on training data was 0.91 and on test data was 0.87.

- XGBOOST
  XGBoost is one of the supervised machine learning algorithms that were used in this project. It is a tree-based algorithm that is an optimized version of the gradient boosting algorithm combined with tree-pruning, parallel processing, and regularization to avoid overfitting or underfitting the model. The accuracy of the XGBoost classifier on training data was 0.99 and on test data was 0.97.

- DECISION TREE
  Decision tree is a tree-based supervised machine learning algorithm. Here we start from the root node for predicting the class label. Whenever we have new data, the values are compared between the new data and the root attribute, and depending on the comparison, we select the branch and compare it with the next node. The accuracy of the Decision Tree classifier on training data was 0.99 and on test data was 0.97.

- RANDOM FOREST
  Random Forest is also a supervised machine learning and tree-based algorithm. The term "Forest" signifies a collection of decision trees. This collection is usually trained with the bagging method. This improves the result when compared with just one decision tree. The accuracy of the Random Forest classifier on training data was 0.99 and on test data was 0.97.

**Table 1.** Results obtained after testing different supervised learning algorithms on the created data set.

| Algorithm | Accuracy on Training data | Accuracy on Testing data |
| --- | --- | --- |
| SVM | 0.96 | 0.74 |
| KNN | 0.91 | 0.87 |
| XGBOOST | 0.99 | 0.97 |
| DECISION TREE | 0.99 | 0.97 |
| RANDOM FOREST | 0.99 | 0.97 |

**Table 2.** A comparison of time taken to train the model based on different supervised learning algorithms on the created data set.

| Algorithm | Time taken to train the model |
| --- | --- |
| SVM | 8.4451 |
| KNN | 0.0169 |
| XGBOOST | 1.5727 |
| DECISION TREE | 0.0248 |

| RANDOM FOREST | 0.8212 |
|---|---|

From Table 1, it can be inferred that tree-based algorithms give the best results with an accuracy of 0.99 in training and 0.97 in testing data. Now, among the three algorithms, decision trees are chosen over XGboost and random forest because of lesser training time [15]. As shown in Table 2, for decision trees, the time taken was 0.0248 whereas for random forest 0.8212, it was and for XGboost it was 1.5727. Thus, decision trees were chosen as the machine learning model for the robot.

2.2.2 Serial communication between Arduino and Raspberry Pi

In this application, the Arduino has to send an array of distance data to the Raspberry Pi, and then listen for commands issued by it. This requires a bi-directional communication. The USB connection between Arduino and Raspberry Pi is utilized for serial flow of data. Raspberry Pi uses the pyserial library provided by python for serial transfer of data.

*2.3 Methodology*

2.3.1 Designing Arduino controlled obstacle avoidance

Before factoring in machine learning, a simple obstacle avoidance algorithm run by Arduino is implemented in Tinkercad [24]. Here the algorithm is tested by providing different test cases and observing the motors' RPM. This is done in order to fine-tune the algorithm.

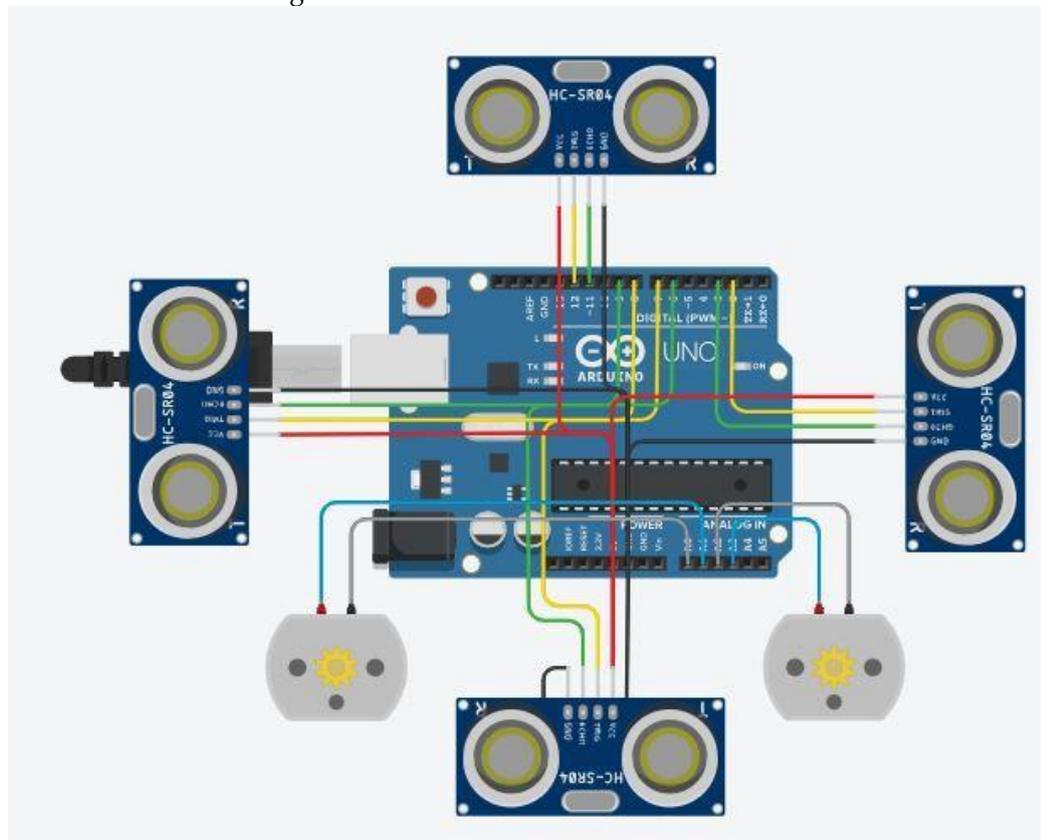

**Figure 2.** Simulating Arduino controlled avoidance in Tinkercad.

The following are the same connections for interfacing the Arduino with the sensors shown in Figure 2 in the actual implementation.

2.3.2 Setting up the Raspberry Pi

The Raspberry Pi must be loaded with an appropriate operating system and necessary libraries must be installed. For the implementation of this project, the latest Raspberry Pi OS is loaded using the Raspberry Pi Imager tool [25] on a micro-SD card no less than 8 GB in capacity. During the setup, provide credentials to the available Wi-Fi network so that when Raspberry Pi boots up for the first time, it connects the said network and is accessible via SSH. PuTTY, an open-source terminal emulator is used for accessing Raspberry Pi's terminal using SSH. For working with Raspberry Pi's desktop, VNC viewer is used. Furthermore, VNC cloud services can be utilized for connecting with the Raspberry Pi without being limited by the range of the same LAN network on which the local PC is connected. The required software libraries can now be installed:

i. $ sudo apt-get update
ii. $ sudo apt-get upgrade
iii. $ sudo apt-get install Arduino
iv. $ pip3 install numpy
v. $ pip3 install pandas
vi. $ pip3 install scikit-learn

2.3.3 Assembling the chassis

The chassis used is a generic 2-WD robot chassis. As the bot needs to house all of the electronics, lightweight plastic-based platforms were created for getting extra space for mounting the components. Holders are also created to secure the components in their place. Ultrasonic sensors are strategically mounted on the front, front-left, front-right and back of the chassis.

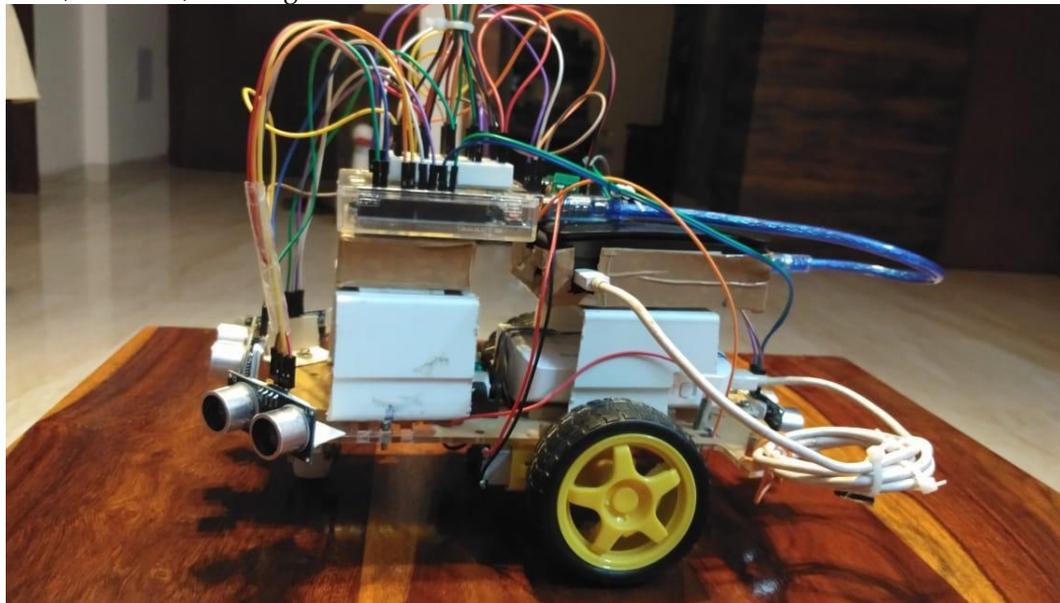

**Figure 3.** Side view of the bot, illustrating the plastic platforms created to mount the Arduino and the Raspberry Pi.

2.3.4 Incorporating machine learning

Employing the obstacle avoidance algorithm run by the Arduino for navigating is a rudimentary approach, and is inadequate for dealing with particular scenarios. In a situation where the robot encounters obstacles that are equidistant from the left and right sensors, the robot starts to oscillate. Therefore, it is observed that the robot needs to continuously learn from previous situations in order to improve performance. To this end, machine learning is introduced. The first task is to create the data set. For this purpose, the bot is deployed and it navigates on the basis of the code running in the Arduino. While it is travelling, distance readings from all four sensors along with the command issued by the Arduino are sent to the Raspberry Pi in the form of an array. The Raspberry Pi stores this data in a comma separated values (csv) file, first few rows of which are shown in Figure 4. The labelled targets are incorrect in situations where the bot gets stuck in some critical situation such as oscillation. These targets are then changed to the correct labels manually to obtain the final training dataset for the machine learning models.

|    | Front  | Back   | Left  | Right | Command |
|----|--------|--------|-------|-------|---------|
| 2  | 128.44 | 82.77  | 81.02 | 74.99 | front   |
| 3  | 127.98 | 196.38 | 80.09 | 74.95 | front   |
| 4  | 127.59 | 84.15  | 79.73 | 74.88 | front   |
| 5  | 128.54 | 82.43  | 79.51 | 66.38 | front   |
| 6  | 128.54 | 83.95  | 79.95 | 63.33 | front   |
| 7  | 126.72 | 72.15  | 79.73 | 74.51 | front   |
| 8  | 125.15 | 84.64  | 78.97 | 74.72 | front   |
| 9  | 126.16 | 84.63  | 78.17 | 71.79 | front   |
| 10 | 124.92 | 84.15  | 78.51 | 74.36 | front   |

**Figure 4.** First few rows of the data set created by the robot.

A machine learning model that is based on decision trees is then created and trained on the collected data. This trained machine learning model is stored inside the Raspberry Pi. During implementation, the Arduino sends an array containing distance values to the Raspberry Pi, which feeds the data to the trained machine learning model and writes back the output as a single byte - 'f', 'b', 'l', 'r', 's' indicating a command to move front, back, left, right, or stop respectively to the Arduino.

2.3.5 Implementation

The user initially inputs a value for threshold distance and critical distance based on their setup. During the time of testing, threshold distance was set as 20cm and critical distance was set as 5cm. As soon as the robot is deployed, the Arduino continuously polls ultrasonic sensors and calculates the distance from obstacles. These distance values are compared with the threshold distance and critical distance specified by the user. If the distance readings from all four sensors are above the critical distance, the Arduino sends the array of distance values to the Raspberry Pi using the pyserial library. The Raspberry Pi uses the machine learning model to calculate a safe move - move front, left, right, back or stop to navigate. In order to implement this safe move, utility functions are defined. To move front, the Arduino sends a HIGH logic to the positive terminal of both the motors and LOW logic is sent to the negative terminal of both the motors. To move left, HIGH logic is sent to the positive terminal of the right motor and negative terminal of the left motor. The remaining terminals receive a LOW logic. This maneuver ensures that the robot rotates in its place without moving forward, which helps to avoid potential crashes by maintaining minimum turning radius. Similar reasoning is used to move the robot right and back. To stop the robot, LOW logic is written on all terminals. Based on the command, the Arduino drives the motors. In case distance reading of any sensor is below the critical distance, it implies that an obstacle is present in very close proximity to the robot and therefore the robot must immediately react to avoid a collision. In such a scenario, it is not feasible to use machine learning based avoidance because it involves transmission latency caused by sending data to the Raspberry Pi and receiving a command back to the Arduino. Hence, if the distance to an obstacle is less than the critical distance, the Arduino itself takes care of avoidance without sending data to the Raspberry Pi. This ensures minimum reaction time to prevent collision with rapidly moving objects. Once the distance reading exceeds the critical distance value, once again the robot uses machine learning to navigate. An overview of this overall flow is shown in figure 5.

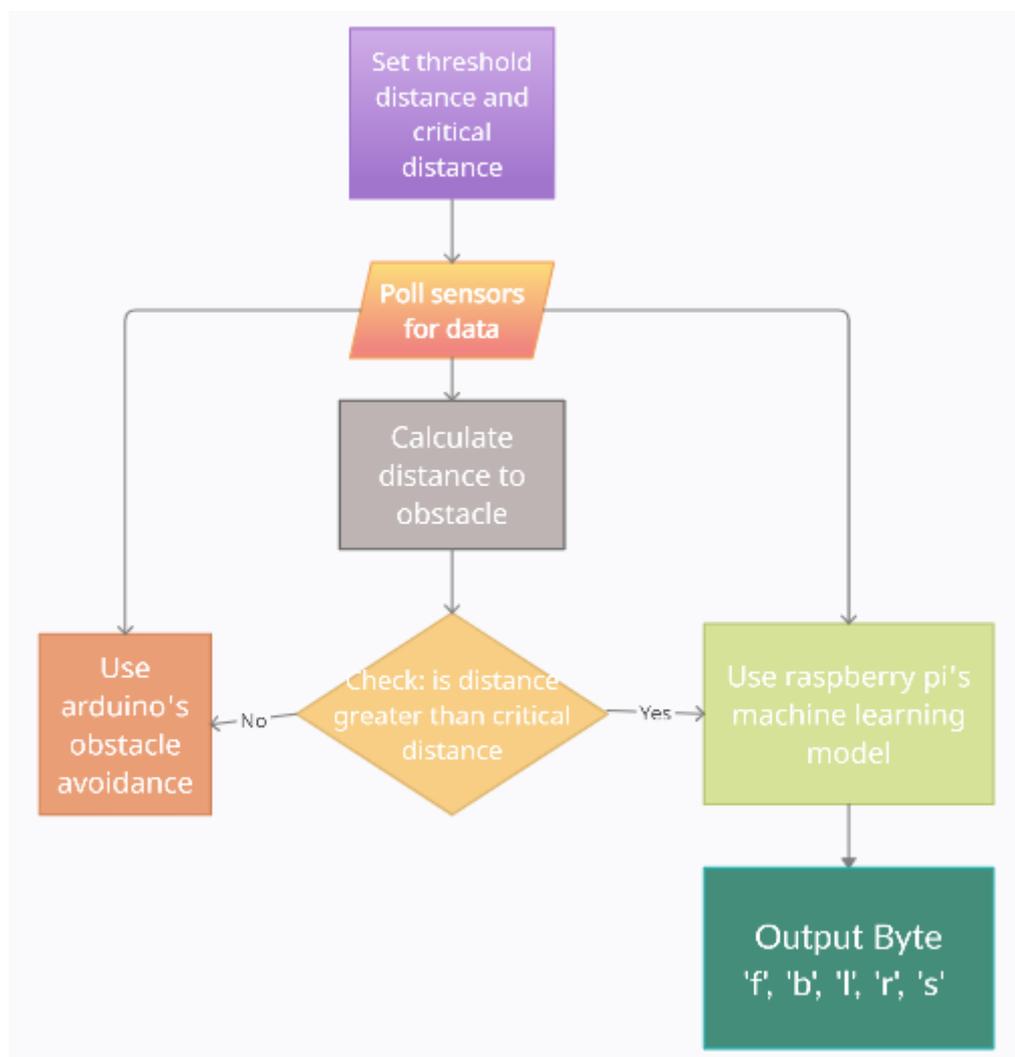

**Figure 5.** Overview of the flow

## 3. Results

The robot was rigorously tested with different test cases including an obstacle course with static objects and a scenario with mobile obstacles such as human beings was also put to the test. The robot was also put in an enclosure that had a small margin for mobility to see if it could find a way out.

*3.1 Obstacle course*

The robot was allowed to move freely in a test course as shown in figure 6, which included stationary objects that were large enough to be detected by ultrasonic sensors. The robot performed admirably, altering its path to move past any objects in its path.

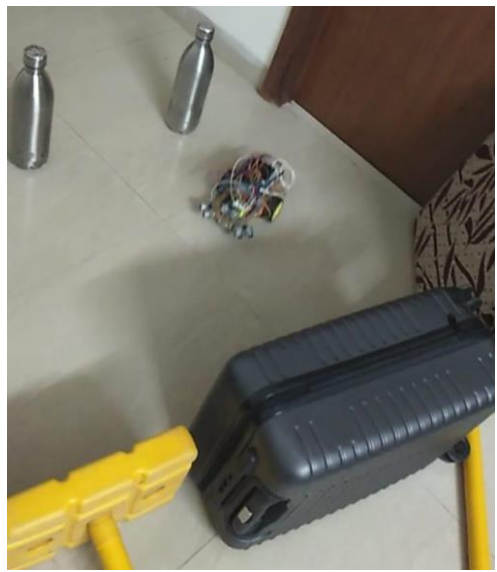

**Figure 6.** Testing the robot in the test course

*3.2 Enclosure test*

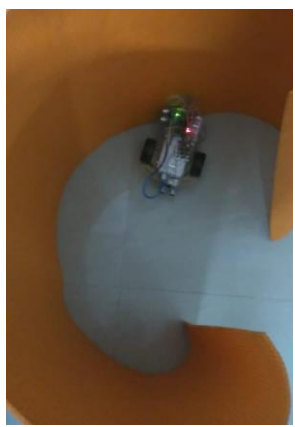 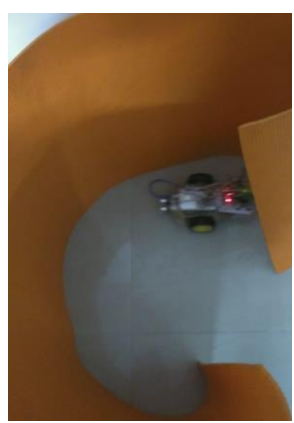

(a) (b)

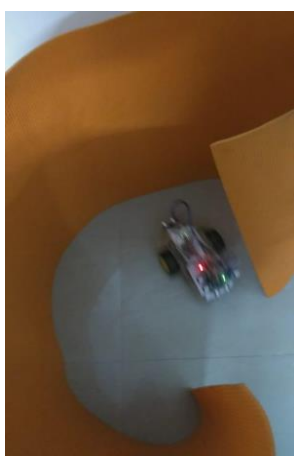 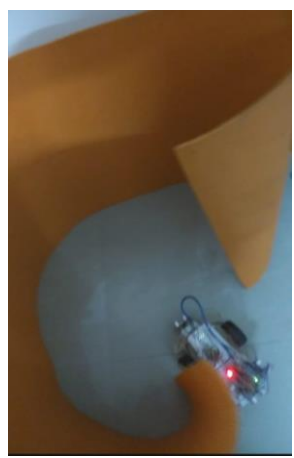

(c) (d)

**Figure 7.** Enclosure test. Initially the bot is kept in the position shown in (a). It then follows the sequence depicted in (b) and (c) followed by a clean exit as shown in (d).

In this test, the robot is placed inside a confined enclosure with a singular opening, with the objective of testing agility and maneuverability. The robot proved to be dexterous enough, and succeeded in escaping through the enclosure without coming in contact with the enclosure walls.

*3.3 Mobile obstacle test*

In this test, objects were placed randomly in close proximity of the robot while it was moving to appraise its reaction time. In all iterations, the robot succeeded in avoiding the object, which shows that the robot has a sufficiently low delay in response.

**4. Discussion**

The robot completed the three tests that it underwent successfully, which proves that the methodology followed is optimal. An argument can be made about including the Arduino at all, since the Raspberry Pi's GPIO pins can be used to interface all the peripherals. However, ultrasonic sensors operate at 5V as opposed to the 3.3V of the GPIO pins. This implies that one needs to build an additional voltage divider circuit for this purpose. Furthermore, circuits for preventing reverse-current must be put in place so that the motor driver's power input from the 6V battery does not affect the Raspberry Pi. Arduino is built for embedded applications and is very easy to interface with sensors as it has an array of robust input/output pins. It was also necessary to implement machine learning because observations were made that the robot sometimes used to get stuck in corners without it. During this case, the bot started to oscillate between left and right since whichever direction it rotated, for that instant of time distance exceeded the threshold in the opposite direction, which prompted the bot to rotate in that direction. This condition continued in a loop and the bot was stuck. After implementing machine learning, this was remedied and the bot was able to execute suitable measures for escaping. It also became apparent that there was a slight delay in receiving commands while using the machine learning model, since it involved a flow of data which brought transmission delay. This was unacceptable for rapidly moving obstacles as it compromised the reaction time. Machine learning is only employed if the distance to the obstacle is greater than the critical distance to solve this problem. This ensures that if an obstacle is dangerously close to the bot, the Arduino calls the shots with minimum reaction time. Although the advantages of using ultrasonic sensors are many, one shortcoming is that dimensionally small obstacles are not detected. Nevertheless, this approach allows for lightweight algorithms which can run efficiently on memory-constrained devices. This also spares computational power that can be utilized to run other tasks like image processing or integrating with ROS for mapping environments.

**5. Conclusions**

This paper proposed and implemented a low-cost and computationally efficient autonomous bot with reactive navigation. Distance readings from four ultrasonic sensors are recorded and a machine learning model based on decision trees was chosen because of its superior accuracy and less training time when compared to other algorithms. After training the model the robot is deployed. By choosing to use machine learning in specific cases, an optimal mix of reaction speed and accuracy is achieved, which enables the robot to move freely in a cluttered indoor set up while being agile enough to avoid mobile obstacles as well.

**Supplementary Materials:** Code repository for this project: https://github.com/yash14s/Jerry-MS

**Author Contributions:** Conceptualization, Y.S., S.S. and S.P.S.; methodology, Y.S. and S.S.; software, Y.S. and S.S.; validation, Y.S. and S.S.; formal analysis, Y.S. and S.S.; investigation, Y.S. and S.S.; resources, Y.S.; data curation, Y.S.; writing—original draft preparation, Y.S. and S.S.; writing—review and editing, S.P.S.; supervision, S.P.S.; All authors have read and agreed to the published version of the manuscript.

**Funding:** This research received no external funding.

**Data Availability Statement:** The data set created by the robot on which the machine learning model is trained on is available at https://github.com/yash14s/Jerry-MS/blob/main/data.csv.

**Acknowledgements:** The authors wish to thank D.P. Srivastava for his assistance in chassis assembly.

**Conflicts of Interest:** The authors declare no conflict of interest.

**Appendix A**

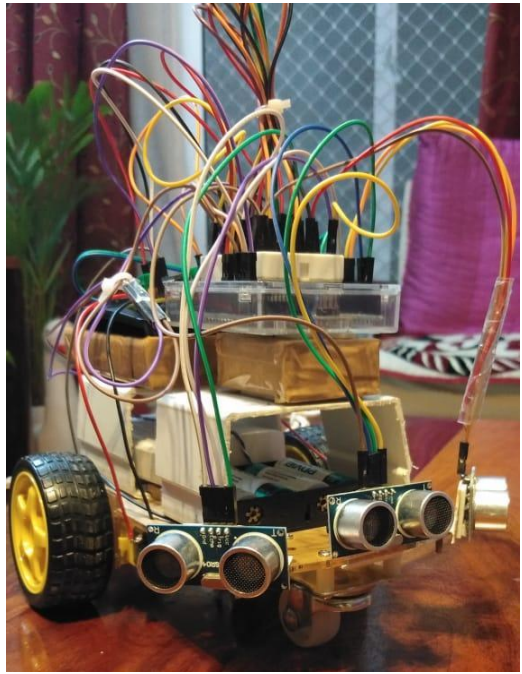

**Figure A1**: Front view of the robot

## References

1. Kitano, H., & Tadokoro, S. (2001). RoboCup Rescue: A Grand Challenge for Multiagent and Intelligent Systems. *AI Magazine*, *22*(1), 39. https://doi.org/10.1609/aimag.v22i1.1542
2. M. Le and M. Le, "Human Detection and Tracking for Autonomous Human-following Quadcopter," 2019 International Conference on System Science and Engineering (ICSSE), 2019, pp. 6-11, doi: 10.1109/ICSSE.2019.8823343.

3. Pierson, H. and Michael S. Gashler. "Deep learning in robotics: a review of recent research." Advanced Robotics 31 (2017): 821 - 835.
4. Yoon, M. et al. "New reinforcement learning algorithm for robot soccer." ORiON 33 (2017): 1-20.
5. S. Islam and A. Razi, "A Path Planning Algorithm for Collective Monitoring Using Autonomous Drones," 2019 53rd Annual Conference on Information Sciences and Systems (CISS), 2019, pp. 1-6, doi: 10.1109/CISS.2019.8693023.
6. Chella, Antonio & Iocchi, Luca & Macaluso, Irene & Nardi, Daniele. (2006). Artificial Intelligence and Robotics. Intelligenza Artificiale. 3. 87-93.
7. Rojas, Eyberth & Rodríguez, Saith & Perez, A.K. & Gomez, Andres & Baez, Heyson & Lopez, J.. (2011). Artificial intelligence system of a robot soccer team. 93-120.
8. J. Wang et al., "Edge-Based Live Video Analytics for Drones," in IEEE Internet Computing, vol. 23, no. 4, pp. 27-34, 1 July-Aug. 2019, doi: 10.1109/MIC.2019.2909713.
9. T. Dash, Soumya Ranjan Sahu, T. Nayak and G. Mishra, "Neural network approach to control wall-following robot navigation," 2014 IEEE International Conference on Advanced Communications, Control and Computing Technologies, 2014, pp. 1072-1076, doi: 10.1109/ICACCCT.2014.7019262.
10. G. Prabhakar, B. Kailath, S. Natarajan and R. Kumar, "Obstacle detection and classification using deep learning for tracking in high-speed autonomous driving," 2017 IEEE Region 10 Symposium (TENSYMP), 2017, pp. 1-6, doi: 10.1109/TENCONSpring.2017.8069972.
11. Freire, Ananda & Barreto, Guilherme & Veloso, Marcus & Varela, Antonio. (2009). Short-term memory mechanisms in neural network learning of robot navigation tasks: A case study. 2009 6th Latin American Robotics Symposium, LARS 2009. 1 - 6. 10.1109/LARS.2009.5418323.
12. M. Zdrodowska, A. Dardzińska and A. Kasperczuk, "Using Data Mining Tools in Wall-Following Robot Navigation Data Set," 2020 International Conference Mechatronic Systems and Materials (MSM), 2020, pp. 1-5, doi: 10.1109/MSM49833.2020.9201730.
13. Thomas, Ajith & Hedley, John. (2019). FumeBot: A Deep Convolutional Neural Network Controlled Robot. Robotics. 8. 62. 10.3390/robotics8030062.
14. Arulkumaran, Kai, Marc Peter Deisenroth, Miles Brundage, and Anil Anthony Bharath. "A brief survey of deep reinforcement learning." *arXiv preprint arXiv:1708.05866* (2017).
15. T R, Prajwala. (2015). A Comparative Study on Decision Tree and Random Forest Using R Tool. IJARCCE. 196-199. 10.17148/IJARCCE.2015.4142.
16. Ma, Runzhuo, Erik B. Vanstrum, Ryan Lee, Jian Chen, and Andrew J. Hung. "Machine learning in the optimization of robotics in the operative field." *Current Opinion in Urology* 30, no. 6 (2020): 808-816.
17. El-Shamouty, Mohamed, Kilian Kleeberger, Arik Lämmle, and Marco Huber. "Simulation-driven machine learning for robotics and automation." *tm-Technisches Messen* 86, no. 11 (2019): 673-684.
18. Mosavi, Amir, and Annamaria R. Varkonyi-Koczy. "Integration of machine learning and optimization for robot learning." In *Recent Global Research and Education: Technological Challenges*, pp. 349-355. Springer, Cham, 2017.
19. Mirza, Nada Masood. "Machine Learning and Soft Robotics." In *2020 21st International Arab Conference on Information Technology (ACIT)*, pp. 1-5. IEEE, 2020.
20. Jain, Aditya Kumar. "Working model of self-driving car using convolutional neural network, Raspberry Pi and Arduino." In *2018 Second International Conference on Electronics, Communication and Aerospace Technology (ICECA)*, pp. 1630-1635. IEEE, 2018.
21. Oltean, Stelian-Emilian. "Mobile robot platform with Arduino uno and raspberry pi for autonomous navigation." *Procedia Manufacturing* 32 (2019): 572-577.
22. Khan, Md Al-Masrur, Md Rashed Jaowad Khan, Abul Tooshil, Niloy Sikder, MA Parvez Mahmud, Abbas Z. Kouzani, and Abdullah-Al Nahid. "A Systematic Review on Reinforcement Learning-Based Robotics Within the Last Decade." *IEEE Access* 8 (2020): 176598-176623.
23. Rossi, Alfa, Nadim Ahmed, Sultanus Salehin, Tashfique Hasnine Choudhury, and Golam Sarowar. "Real-time Lane detection and Motion Planning in Raspberry Pi and Arduino for an Autonomous Vehicle Prototype." *arXiv preprint arXiv:2009.09391* (2020).
24. https://www.tinkercad.com/
25. https://www.raspberrypi.org/software/